%% file: paper.tex
\documentclass{article} 
\usepackage{iclr2020_conference,times}

\input{math_commands.tex}

\usepackage{hyperref}
\usepackage{url}

\usepackage[shortcuts]{extdash}
\usepackage{graphicx} 
\usepackage{mathtools}
\usepackage{amssymb}
\usepackage[skins,breakable]{tcolorbox}
\usepackage{wrapfig,booktabs}
\usepackage{subcaption}

\title{Affine Self Convolution}

\author{Nichita Diaconu \& Daniel Worrall \\
Philips lab\\
University of Amsterdam\\
\texttt{diacon995@gmail.com,d.e.worrall@uva.nl}}

\iclrfinalcopy 
\begin{document}
\maketitle
\begin{abstract}
Attention mechanisms, and most prominently self\-/attention, are a powerful building block for processing not only text but also images. These provide a parameter efficient method for aggregating inputs. We focus on self\-/attention in vision models, and we combine it with convolution, which as far as we know, are the first to do. What emerges is a convolution with data dependent filters. We call this an Affine Self Convolution. While this is applied differently at each spatial location, we show that it is translation equivariant. We also modify the Squeeze and Excitation variant of attention, extending both variants of attention to the roto-translation group. We evaluate these new models on CIFAR10 and CIFAR100 and show an improvement in the number of parameters, while reaching comparable or higher accuracy at test time against self\-/trained baselines.
\end{abstract}
\section{Introduction}
Computer vision has seen great success thanks to the use of the convolution operation in Convolutional Neural Networks (CNNs) \citep{Lecun89,Krizhevsky12,He17MaskRCNN,Mnih13}. This operation takes advantage of the translational symmetry in visual perception tasks such as image classification. Meanwhile, in tasks that require sequence processing, attention \citep{chorowski2015attention,Bahdanau14} and self\-/attention \citep{Vaswani17} have emerged as a powerful technique.

One of the peculiarities of CNNs is that filters are defined independently of the data. At the same time, self\-/attention is data dependent, but does not provide a template matching scheme, as does the convolution operation, since it merely reweights neighborhoods. While there is work towards using attention in CNNs, the current models use them independently, sequentially, or in parallel. 

We unify convolution and self\-/attention, taking the best of both worlds. Our method provides a translationally equivariant convolution, where the filters are also dependent on the input. These data dependent filters more efficiently describe the relations present in the input. Moreover, by formulating it as a special convolution, it can be extended to be equivariant to other groups of transformations. As a result, we apply the rich literature on group equivariant neural networks \citep{CohenW16} and develop the roto-translation equivariant counterpart. This module can be used as a replacement for standard convolutional layers and we call it an Affine Self Convolution.

Another variant of attention in computer vision is Squeeze and Excitation \citep{SE}. This provides global attention and we extend it to the roto-translation variant in order to compare it to our module. We plan to release code for all the experiments soon.

The contributions of this work are:
\begin{itemize}
    \item[$\bullet$] We introduce the Affine Self Convolution (ASC), merging convolution and self\-/attention.
    \item[$\bullet$] We prove ASC is translation equivariant.
    \item[$\bullet$] We extend ASC to roto\-/translation equivariant ASC. 
    \item[$\bullet$] We develop group Squeeze and Excitation. 
    \item[$\bullet$] We evaluate these modules on CIFAR10 and CIFAR100.
\end{itemize}
\section{Background}
In order to combine convolution and self\-/attention we first look at the group convolutions \citep{cohen2018general} and then at the self\-/attention mechanism \citep{Vaswani17,parmar2018imagetransformer}.
\subsection{Group convolution}
Group equivariant convolutional neural networks extend the operation of convolution. \citet{CohenW16} show that we can consider the convolution operation to be defined on a group and that this allows for a natural generalization to other objects that have a group structure.

\textbf{Translation equivariant convolution} The set of points in \(\mathbb{Z}^2\) with the operation of vector addition forms a group. For each value \(x\) in this group, the translation operator \(\mathcal{L}_{(x)}\) translates the domain of a function: \(\mathcal{L}_{(x)}[f](y) = f(-x+y)\). Therefore, given two functions \(f,\psi:\mathbb{Z}^2\rightarrow \mathbb{R}\), the convolution between the image \(f\) and the filter \(\psi\) can be written in terms of translations by elements of the group \((\mathbb{Z}^2,+)\) (we depict the convolution in Figure \ref{fig:Z2_conv}):
\begin{align}
[f \star_{\mathbb{Z}^2} \psi](x) &= \sum_{y\in \mathbb{Z}^2}f(y)\mathcal{L}_{(x)}[\psi](y) \label{eqn:Z2_conv}
\end{align}Where the convolution operation \(\star_{\mathbb{Z}^2}\) is indexed by the domain of the input and the filter. Most importantly, the convolution is equivariant to translations: \(\mathcal{L}_{(z)}[f] \star_{\mathbb{Z}^2} \psi = \mathcal{L}_{(z)}[f \star_{\mathbb{Z}^2} \psi ]\). This property connects a transformation of the input image \(\mathcal{L}_{(z)}[f]\) with a precise transformation of the activations \(\mathcal{L}_{(z)}[f \star_{\mathbb{Z}^2} \psi ]\). This is desirable because a model that implements such an operation can also be invariant to translations by taking a max over spatial positions at the end.

\textbf{Roto-translation equivariant convolution} A natural extension of the group of translations is the group of planar rotations and translations \citep{weiler2018learning,diaconu2019learning}. The elements of this group have 2 components, a (proper) rotation \(R\) and a translation \(y\):
\begin{align}
    SE(2)&=\{(R,y)\in(R\in \mathbb{R}^{2\times 2}, y\in \mathbb{R}^{2})|R^\top R=RR^\top = I , \text{det}(R)=1)\}
\end{align}Where \(I\) is the identity matrix of order 2 and \((I,0)\) is the identity element of \(SE(2)\). Throughout this work we will use \(P,R,S\) to denote rotation elements and \(x,y,z\) translation elements of the group \(SE(2)\). From here, we denote the \(G=SE(2)\). Similarly to \citet{diaconu2019learning}, the operator \(\mathcal{L}_{(P,x)}\) inversely transforms the coordinates of a function \(f\) with domain \(\mathbb{R}^2\) by \(\mathcal{L}_{(P,x)}[f](y) = f(P^{-1}(y-x))\). As a result, we can evaluate the planar convolution at points on \(G\), \([f \star_{\mathbb{R}^2} \psi](P,x)\). This operation is called a lifting layer and outputs activations with domain \(G\). The operator \(\mathcal{L}_{(P,x)}\) transforms such functions by \(\mathcal{L}_{(P,x)}[f](R,y) = f(P^{-1}R,P^{-1}(y-x))\). We can preserve \(G\) equivariance of these functions by replacing the planar convolution with \(G\) convolutions between \(f,\psi:G\rightarrow \mathbb{R}\):
\begin{align}
[f \star_{G} \psi](P,x) &= \sum_{(R,y)\in G}f(R,y)\mathcal{L}_{(P,x)}[\psi](R,y) \label{eqn:G-conv}
\end{align}In practice we are limited by the sampling grid, as real world images are functions on \(\mathbb{Z}^2\). This has as symmetries translations by integer values, which form the group \(\mathbb{Z}^2\) and rotations by multiples of \(90^\circ\), which form the group \(C_4\). By replacing the groups \(\mathbb{R}^2\) and \(SO(2)\) in the definition of \(SE(2)\) with \(\mathbb{Z}^2\) and \(C_4\) we get the group \(p4\). This is a subgroup of \(SE(2)\), and all the relations in this work hold for both.
\subsection{Self-attention in Computer vision} \begin{figure}[t]
  \centering
  \begin{minipage}[b]{0.49\textwidth}
    \includegraphics[width=1\linewidth]{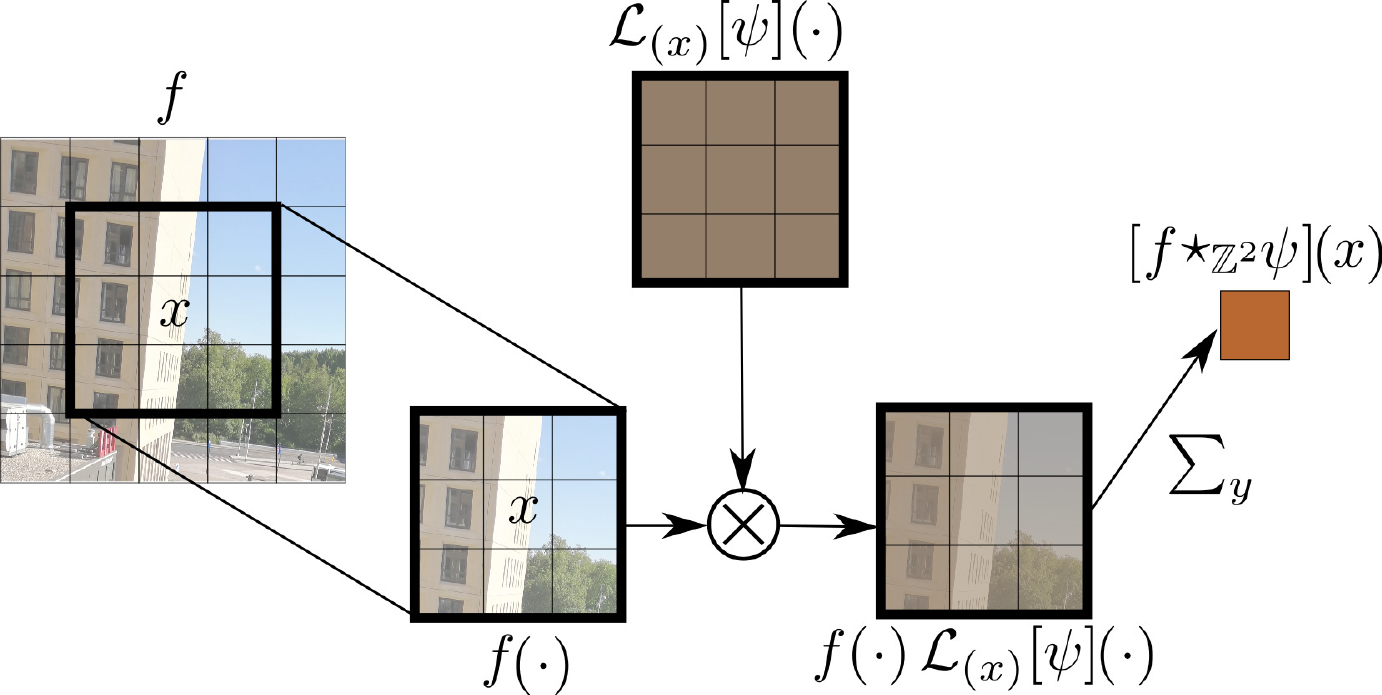}
    \caption[Convolution]{Convolution between an input \(f\) and a \(3\)\(\times\)\(3\) filter \(\psi\) evaluated at \(x\). We denote with \((\cdot)\) all the positions \(y\) in the neighborhood of \(x\). This begins by cropping the \(3\)\(\times\)\(3\) neighborhood in \(f\) centered at \(x\). The filter is colored brown for clarity, but it can have different values. The filter is centered in the same neighborhood by translating it \(\mathcal{L}_{(x)}[\psi](\cdot)\). The two quantities are then multiplied at each spatial position \(y\), independently. Finally, the sum over the spatial positions aggregates the response of the convolution, concluding how similar the input neighborhood around \(x\) is to the filter.} \label{fig:Z2_conv}
  \end{minipage}
  \hfill
  \begin{minipage}[b]{0.49\textwidth}
    \includegraphics[width=1\linewidth]{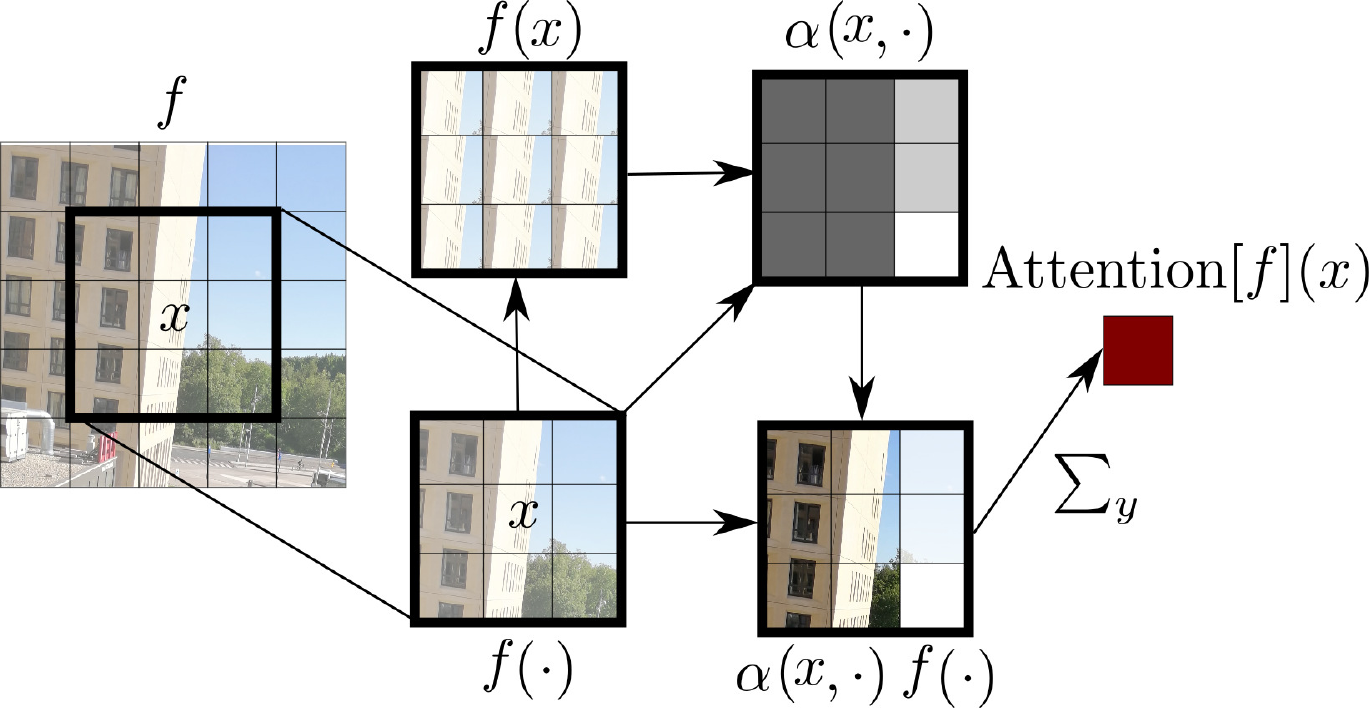}
    \caption[Simple self\-/attention]{Simple self\-/attention with extent \(3\)\(\times\)\(3\) for an input \(f\) evaluated at \(x\). Similarly to the convolution \((\cdot)\) denotes the neighborhood of \(x\). The \(3\)\(\times\)\(3\) region around \(x\) is cropped. The score function then compares the center with each neighbor. The normalized score is \(\alpha(x,y)\). The scores and the input image neighborhood are then multiplied at each spatial position. This is shown over \(\alpha(x,\cdot)f(\cdot)\) by sharpening the color on the building and whitening or removing the color from the sky or trees. By summing over the spatial dimension, we aggregate the neighborhood scaled by how similar the center is to its neighbors.}\label{fig:Z2_self_attn}
  \end{minipage}
\end{figure} \textbf{Simplified form} We start with a simplified form of local self\-/attention to highlight each part and to make notation clearer. For an input image \(f:\mathbb{Z}^2\rightarrow\mathbb{R}\),
self\-/attention is defined by 2 parts: 1) a score function between center pixel \(f(x)\) and a neighbor pixel \(f(y)\). This is usually the dot product, \(\text{score}(f(x),f(y)) = f(x)f(y)\). Moreover, such that the score values in a neighborhood sum to \(1\), the scores are normalized with softmax, \(\alpha(x,y) = \text{softmax}_y(\text{score}(f(x), f(y))\), 2) an aggregation of the neighbors \(y\) based on the score function: 
\begin{align}
\text{Attention}[f](x) &= \sum_{y\in\mathbb{Z}^2} \alpha(x,y)f(y) \label{eqn:Z2_simple_selfattn}
\end{align}This operation defines a neighborhood dependent weighting. We depict this in Figure \ref{fig:Z2_self_attn}. The dot product score function, and therefore Equation \ref{eqn:Z2_simple_selfattn}, does not take account of the relative position between \(x\) and \(y\). Simply, this encodes no spatial information. To take advantage of the regular grid in images, recent methods use positional embeddings \(\beta:\mathbb{Z}^2\rightarrow \mathbb{R}\). These can be added to neighbors at \(y\), prior to computing the score, and is parametrized by the relative position from the center \(x\), as done in \citet{ramachandran2019SASA}, \(\alpha(x,y) = \text{softmax}_y(\text{score}(f(x),f(y)+\beta(y-x))) \label{eqn:pos_emb_betaK}\). It can also be added to the neighbors score, after computing the score function, as done in \citet{bello2019AACNN,hu2019LRNet}, \(\alpha(x,y) = \text{softmax}_y(\text{score}(f(x), f(y))+\beta(y-x))\)

\textbf{Self-attention} In practice, self\-/attention uses three sets of parameters \citep{parmar2018imagetransformer,wang2018NLNet}, \(W^Q\), \(W^K\), \(W^V\), where \(Q,K,V\) stands for Query, Key, Value. We turn our attention to an input function (image) \(f:\mathbb{Z}^2\rightarrow\mathbb{R}^{d_{\text{in}}}\), with \(d_{\text{in}}\) input channels and parameters \(W^Q,W^K:\mathbb{Z}^2\rightarrow\mathbb{R}^{d_{k}\times d_{\text{in}}}\text{ and } W^V:\mathbb{Z}^2\rightarrow\mathbb{R}^{d_{\text{out}}\times d_{\text{in}}}\). \(W^Q\), \(W^K\), \(W^V\), which are implemented as \(1\)\(\times\)\(1\) convolutions, have the purpose of mapping to three separate embeddings, \(Q\), \(K\), \(V\) and not to process spatial information. These have \(d_k\) or \(d_{\text{out}}\) channels. The spatial information is weighted and mixed by the attention mechanism. In this more general setting, where we denote an arbitrary channel with \(c\), self\-/attention is defined with: 1) three linear mappings of the input \(f\) (defined analogously for \(K\) and \(V\)): \(Q_c(x) =[f\star_{\mathbb{Z}^2}W^Q_c](x)\) 2) a normalized score function, which can use positional embeddings: \(\alpha(x,y) = \text{softmax}_y(\text{score}(Q(x), K(y)))\) and 3) an aggregation of the \(V\) embeddings based on the score function:
\begin{align}
    \text{Attention}[Q,K,V]_c(x) &= \sum_{y\in\mathbb{Z}^2} \alpha(x,y)V_c(y) \label{eqn:Z2_self_attn}
\end{align}It is also common to use the multi head mechanism from Transformer \citep{Vaswani17} alongside attention. This means that \(Q,K,V\) are split along the channel dimension and the self\-/attention mechanism is evaluated independently for each element of the partition (each head). These are then concatenated and passed through a linear layer.
\section{Method}
In the following section we look at the local relative type of attention, self\-/attention and we formulate it similarly to convolution. This allows us to merge the two and then develop the roto-translational variant. We also derive the roto-translational variant of the Squeeze and Excite module, which defines global attention.

\subsection{Affine Self Convolution (ASC)} \label{sec:ASC}
\begin{figure}[t]
\includegraphics[scale=0.54]{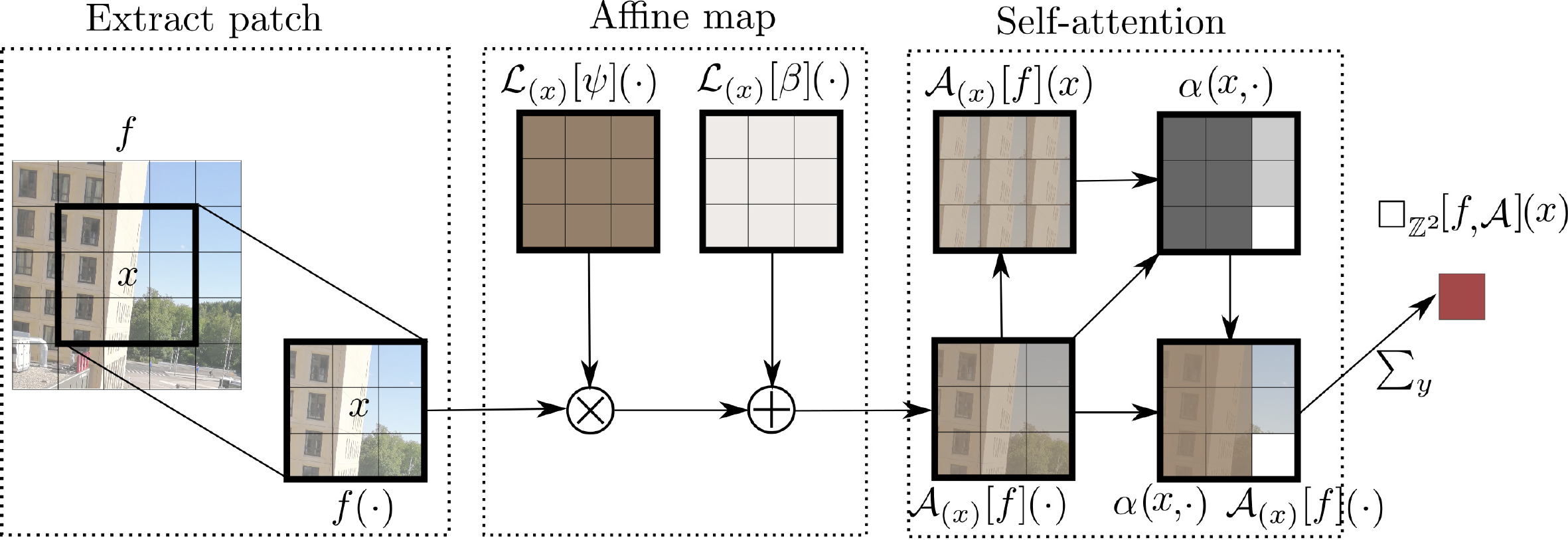}
\centering
\caption[Simple Affine Self Convolution]{Simple ASC for an input \(f\) and an affine map \(\mathcal{A}\) with extent \(3\)\(\times\)\(3\) evaluated at \(x\). As for Figures \ref{fig:Z2_conv} and \ref{fig:Z2_self_attn}, \((\cdot)\) denotes the neighborhood of \(x\). The neighborhood is first extracted from the input, then the affine map parameters are centered in the neighborhood of \(x\). For each spatial position, \(f\) is transformed affinely by \(\mathcal{L}_{(x)}[\psi](\cdot)\) and \(\mathcal{L}_{(x)}[\beta](\cdot)\), which results in \(\mathcal{A}_{(x)}[f](\cdot)\). If we were to sum now over spatial positions, this would be an affine convolution. Nonetheless, the affinely transformed input is now passed through the self\-/attention mechanism. This consists of evaluating a similarity score between the center of the neighborhood and the neighbors \(\alpha(x,y)\) and then multiplying the input with the score function. As a result, the ASC module can focus on the parts of the image that contain information relevant to the center. Similarly to both previous methods, the neighbors' information is aggregated as the final step. Intuitively, what we get is a data dependent convolution between for the image around \(x\).} \label{fig:Z2_Simple_ASC}
\end{figure}
\textbf{Affine map} We have seen how simplified self\-/attention in Equation \ref{eqn:Z2_simple_selfattn} is applied to an input image and that it was further developed using relative positional embeddings \(\beta\) in order to include spatial information. We note that these are additive terms and extend them by a multiplicative term \(\psi\) that is also spatially dependent. This results in a affine map \(\mathcal{A}\), which depends on \(\beta,\psi:\mathbb{Z}^2\rightarrow \mathbb{R}\). We define the map \(\mathcal{A}\) relative to a center \(x\): \(\mathcal{A}_{(x)}:(\mathbb{Z}^2,\mathbb{R})\rightarrow\mathbb{R}\) and we apply it using the translation operator \(\mathcal{L}_{(x)}\) as \(\mathcal{A}_{(x)}(y,r) = r\mathcal{L}_{(x)}[\psi](y)+\mathcal{L}_{(x)}[\beta](y)\). Furthermore, we can describe the affine map as acting on an image \(f:\mathbb{Z}^2\rightarrow\mathbb{R}\):
\begin{align}
    \mathcal{A}_{(x)}[f](y) &\coloneqq \mathcal{A}_{(x)}(y,f(y)) = f(y)\mathcal{L}_{(x)}[\psi](y)+\mathcal{L}_{(x)}[\beta](y) \label{eqn:Z2_affine_map} 
\end{align}

\textbf{Simplified form} Applying the affine map \(\mathcal{A}\), then simplified self\-/attention from Equation \ref{eqn:Z2_simple_selfattn} to \(f\): \(\text{Attention}[\mathcal{A}_{(x)}[f]](x)=\sum_{y} \alpha(x,y)\mathcal{A}_{(x)}[f](y)\) is how we define the simplified ASC. We denote this with \(\square_{\mathbb{Z}^2}\). This is applied to two functions, an input and an affine filter, and we index it by the domain of the two functions, similarly to the convolution operation:
\begin{align}
    \square_{\mathbb{Z}^2}[f,\mathcal{A}](x)&\coloneqq \sum_{y} \alpha(x,y)\mathcal{A}_{(x)}[f](y)= \sum_{y\in\mathbb{Z}^2}\underbrace{\alpha(x,y)}_{\text{score}}( f(y)\underbrace{\mathcal{L}_{(x)}[\psi](y)}_{\text{filter}}+\underbrace{\mathcal{L}_{(x)}[\beta](y)}_{\text{positional emb}}) \label{eqn:Z2_simple_selfattn_expanded}
\end{align} This uses the self\-/attention score from Equation \ref{eqn:Z2_simple_selfattn}, the convolutional filter from Equation \ref{eqn:Z2_conv} and also adds positional embeddings. Moreover, we can distribute \(\alpha\) and view \(\alpha(x,y)\mathcal{L}_{(x)}[\psi](y)\) and \(\alpha(x,y)\mathcal{L}_{(x)}[\beta](y)\) as the parameters of a normalized affine map. Intuitively, this not only performs template matching through \(\psi\), which is independent of the information in the image, but scales the template relative to what is in the image through \(\alpha(x,y)\). By unifying the convolution and self\-/attention, these data dependent filters can more efficiently describe the relations in the image because they are applied differently at each location. We call this an Affine Self Convolution and we depict it in Figure \ref{fig:Z2_Simple_ASC}. It is possible to recover the usual convolution by setting the scaling coefficients \(\alpha(x,y)\) to \(1\) and \(\beta\) to \(0\). By setting \(\psi\) to 1, we recover a simplified self\-/attention, where positional embeddings are used for computing the score and for the aggregated term.

\textbf{Translation equivariance} We prove that this is translation equivariant in the Appendix \ref{eqn:Z2_ASC_proof} and therefore: \(\square_{\mathbb{Z}^2}[\mathcal{L}_{(z)}[f],\mathcal{A}](x) = \mathcal{L}_{(z)}[\square_{\mathbb{Z}^2}[f,\mathcal{A}]](x)\). This means that this model can detect objects regardless of their position in the image, as does the standard convolution.

\textbf{Affine Self Convolution}
Similarly to the general form of self\-/attention in Equation \ref{eqn:Z2_self_attn}, we use three sets of parameters, \((W^Q, \mathcal{A}^Q)\), \((W^K, \mathcal{A}^K)\), \((W^V, \mathcal{A}^V)\) for an input \(f:\mathbb{Z}^2\rightarrow\mathbb{R}^{d_{\text{in}}}\). This is depicted in Appendix Figure \ref{fig:ASC}. By contrast to the simple variant, we now also have the affine maps. These process spatial information and are \(0\) outside the extent, which is controlled by a hyperparameter: kernel size. Moreover, each affine map, can be implemented with \(\psi^Q,\psi^K:\mathbb{Z}^2\rightarrow\mathbb{R}^{d_{k}\times d_{k}}\), mapping from all the channels in the input to all the channels in the output or \(\psi^Q,\psi^K:\mathbb{Z}^2\rightarrow\mathbb{R}^{d_{k}}\), mapping from one channel in the input to the same channel in the output. The same is true for \(\psi^V\), if we replace \(d_k\) with \(d_{\text{out}}\). Our experiments use the latter, due to computational constraints. The additive term of the map is defined as \(\beta^Q, \beta^K:\mathbb{Z}^2\rightarrow\mathbb{R}^{d_{k}}\), where for \(\beta^V\), \(d_k\) is replaced with \(d_{\text{out}}\). We denote an arbitrary channel with \(c\) and we define ASC as:
\begin{align}
\shortintertext{1) three linear mappings of the input \(f\) (defined analogously for \(K\) and \(V\)):}
    Q_c(x) &= [f\star_{\mathbb{Z}^2}W^Q_c](x)
\shortintertext{2) three affine maps \(\mathcal{A}^Q, \mathcal{A}^K, \mathcal{A}^V\) for the \(Q,K,V\) terms (defined analogously for \(K\) and \(V\)):}
    \mathcal{A}^Q_{(x)}[Q]_c(y) &= Q_c(y)\mathcal{L}_{(x)}[\psi^Q]_c(y)+\mathcal{L}_{(x)}[\beta^Q]_c(y) \\
\shortintertext{3) a score function between center \(x\) and neighbor \(y\), which is then normalized with softmax:}
    \alpha(x,y) &= \text{softmax}_y(\text{score}(\mathcal{A}_{(x)}^Q[Q](x), \mathcal{A}_{(x)}^K[K](y)))
\shortintertext{4) an aggregation of the \(V\) embeddings based on the score function:}
    \square_{\mathbb{Z}^2}[V,\mathcal{A}^V]_c(x) &= \sum_{y\in\mathbb{Z}^2} \alpha(x,y)\mathcal{A}_{(x)}^V[V]_c(y) \label{eqn:Z2_ASC_model}
\end{align}We also use the multi head mechanism. We note that now we have a separate set of parameters for \(\mathcal{A}^Q, \mathcal{A}^K, \mathcal{A}^V\), and that \(\mathcal{A}^Q\) is only evaluated at:\(\mathcal{A}^Q_{(x)}[Q](x) = Q(x)\psi^Q+\beta^Q\).
Therefore, we can learn \(\psi^Q, \beta^Q\) for only one spatial index \(0\).

By comparison, the positional embeddings in \citet{ramachandran2019SASA} are represented by the term \(\beta^K\). The positional embeddings in \citet{hu2019LRNet} are equivalent to learning the product of the embeddings \(\beta^Q\beta^K\), which arises when multiplying \(\mathcal{A}^Q_{x}[Q](x)\mathcal{A}^K_{x}[K](y)\). A difference between this work and their work is that we directly learn these parameters, while \citet{ramachandran2019SASA,hu2019LRNet} use a separate network to learn \(\beta\).

\subsection{Roto-translation Affine Self Convolution}
We now use the machinery of this new operation and the group theoretic background to develop the roto-translation ASC. Similarly to the standard ASC, we first define a simplified form based on an affine map, then we prove roto-translation equivariance, and finally, we present the general form.

\textbf{Affine map} We now turn to functions on \(SE(2)\). We will denote \(G\coloneqq SE(2), H\coloneqq SO(2), \mathbb{R}^2=G/H\). Similarly to the relative affine map for functions defined on \(\mathbb{Z}^2\) in equation \ref{eqn:Z2_affine_map}, we define a relative affine map \(\mathcal{A}\) for functions on \(G\). This map uses affine parameters \(\psi,\beta:G\rightarrow\mathbb{R}\) and is defined as \(\mathcal{A}_{(P,x)}:(G,\mathbb{R})\rightarrow\mathbb{R}\). This acts on a function \(f\) with domain \(G\) as:
\begin{align}
    \mathcal{A}_{(P,x)}[f](R,y) &\coloneqq f(R,y)\mathcal{L}_{(P,x)}[\psi](R,y)+\mathcal{L}_{(P,x)}[\beta](R,y)
\end{align}This affine map transforms a function \(f\) relative to a center \((P,x)\).

\textbf{Simplified form} To define the score function, we notice that we can split the sum over the group \(G\) in Equation \ref{eqn:G-conv} into two sums, \(\sum_{y\in G/H}\sum_{R\in H}\mathcal{A}_{(P,x)}[f](R,y)\). In this form, for each \(P\in H\) the convolution can be seen as a weighted sum of the neighbors \(y\), relative to a center \(x\). We can scale the affine group convolution based on this intuition. Precisely, we add a score function that is evaluated between each center \((P,x)\) and each neighbor \(y\):
\begin{align}
    \alpha((P,x),(P,y)) = \text{softmax}_y \left ( \left ( \sum_{R\in H} \mathcal{A}_{(P,x)}[f](R,x)\right ) \left ( \sum_{R\in H} \mathcal{A}_{(P,x)}[f](R,y)\right ) \right ) \label{eqn:p4_score_fn}
\end{align}
\begin{wrapfigure}[10]{r}{0.32\textwidth}
    \centering
    \includegraphics[scale=0.25]{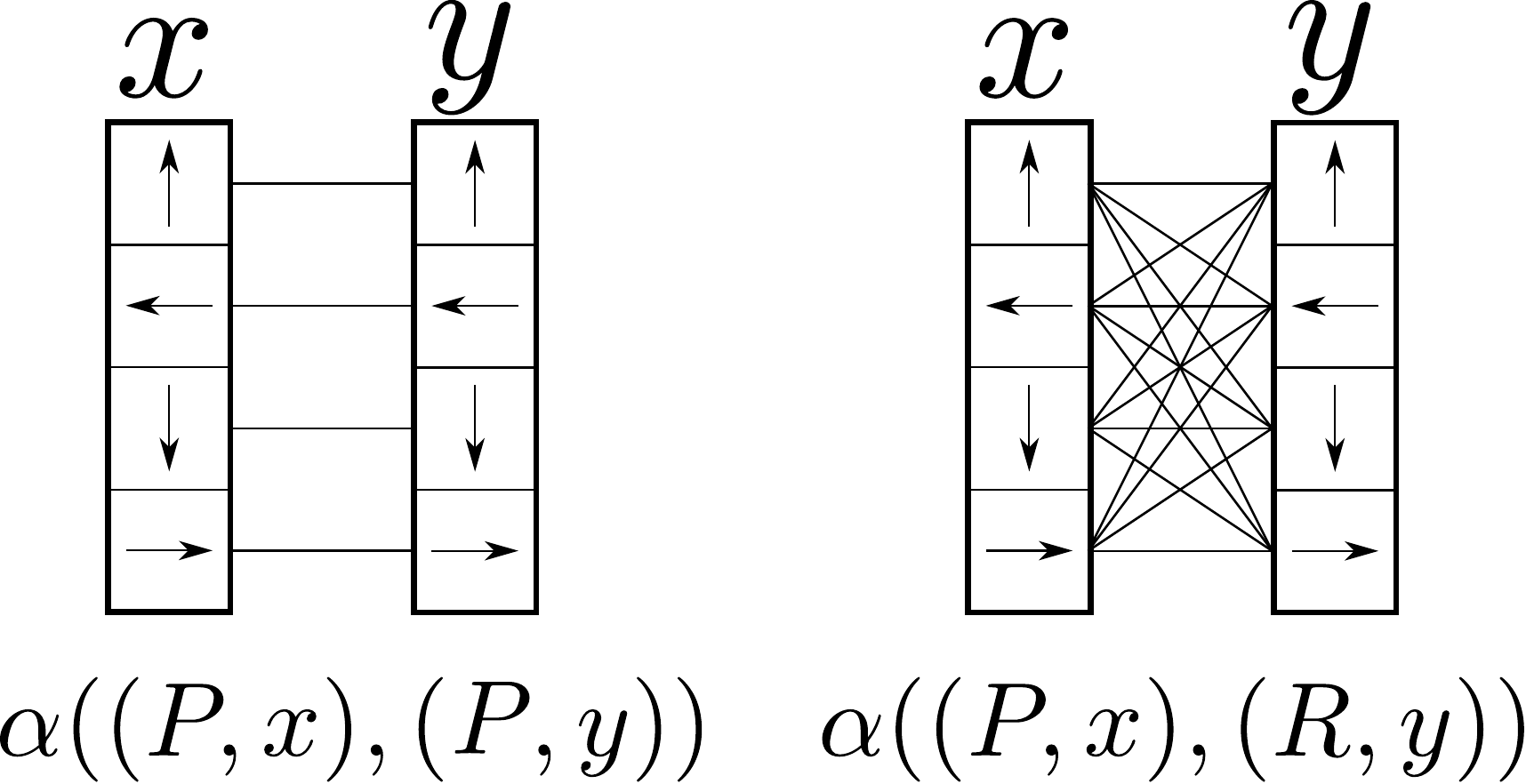}
    \caption{We show the plausible score functions between the equivariant stack of activations at the center \(x\) and the neighbor \(y\). Our models uses the left variant, preserving equivariance.}
    \label{fig:group_score_function}
\end{wrapfigure}
We note that it might also be possible to define a score function \(\alpha((P,x),(R,y))\) and we depict the difference in Figure \ref{fig:group_score_function}. Nonetheless, this could be computationally prohibitive and is not required in order to preserve roto-translation equivariance. Using the score in Equation \ref{eqn:p4_score_fn} we define the Simple ASC on \(G\):
\begin{align}
    \square_G[f,\mathcal{A}](P,x) 
    &=\sum_{y\in G/H} \alpha((P,x),(P,y)) \sum_{R\in H} \mathcal{A}_{(P,x)}[f](R,y) \label{eqn:p4_Simple_ASC}
\end{align}In the Appendix \ref{eqn:p4_affine_map_Z2_beta} we show that we can replace \(\sum_{R\in H}\beta(R,x)\) with \(\beta(x)\). Therefore, we can learn \(\beta\) as a function with domain \(G/H\). We depict this mechanism in the Appendix Figure \ref{fig:Group_Simple_ASC}.

\textbf{Roto-translation equivariance} We verify that ASC on groups is equivariant to actions of the group \(G\), by checking the equivariance relation in the Appendix \ref{eqn:p4_ASC_equivariance}.

\textbf{Roto-translation ASC} In practice, the input is discrete and we turn to functions on \(p4\), \(G\coloneqq p4, H\coloneqq C_4, \mathbb{Z}^2=G/H\). For the general roto-translation ASC all quantities are defined analogously to the ASC on \(\mathbb{Z}^2\) in Equation \ref{eqn:Z2_ASC_model}, using the score function in Equation \ref{eqn:p4_score_fn}, the aggregation in Equation \ref{eqn:p4_Simple_ASC} and replacing the domain \(\mathbb{Z}^2\) with \(G\). The details can be found in Appendix \ref{sec:p4_ASC}. \subsection{Group Squeeze and Excite}
Interactions between filters that are spatially far apart is also tackled in Squeeze and Excitation (SE) \citep{SE}. The SE module proposes to rescale each feature map based on a global aggregation of the spatial dimension. An intuition for this is that, by allowing for channel interactions at all the spatial locations, this effectively enlarges the receptive field maximally. This is also a parameter efficient method for increasing the receptive field. 

For a general group \(G\) (with \(g,h,s\in G\)), the squeeze term takes an average of a function \(f: G\rightarrow \mathbb{R}^{d_{\text{in}}}\) over the group. This is invariant to transformations by actions of the group \(G\), which we show in the Appendix \ref{eqn:squeeze_g_invariant}. The average is then passed through a one hidden layer MLP (\(W_1 \in \mathbb{R}^{d_{\text{in}}\times (d_{\text{in}}/r)}, W_2 \in \mathbb{R}^{(d_{\text{in}}/r)\times d_{\text{in}}}\), where \(d_{\text{in}}\) is the number of channels of the input and \(r\) is the reduction ratio). A sigmoid unit (\(\sigma\)) is then used on the activation. 
\begin{align}
    \text{squeeze}(f) &= \sigma \left (W_1 \left( \text{ReLU} \left ( W_2 \left (\frac{1}{|G|}\sum_{h \in G}f(h) \right ) \right )\right)\right)
\end{align}These are then broadcasted across the domain with an element\-/wise multiplication, \(f'_{c_{\text{in}}}(g)=f_{c_{\text{in}}}(g)\text{squeeze}_{c_{\text{in}}}(f)\).
Therefore, multiplying a function \(f(g)\) by \(\text{squeeze}(f)\) preserves the group structure of \(f\):
\begin{align}
    \mathcal{L}_{(s)}[f']_{c_{\text{in}}}(g)&=\mathcal{L}_{(s)}[f_{c_{\text{in}}}(g)\text{squeeze}_{c_{\text{in}}}(f)] =\mathcal{L}_{(s)}[f]_{c_{\text{in}}}(g)\text{squeeze}_{c_{\text{in}}}(f)
\end{align}This is added as the last operation in any bottleneck ResNet. In our experiments, the group is either the group of integer translations \(\mathbb{Z}^2\) (which is the original operation is \cite{SE} and for which \(|G|\) is the height\(\times\)width of the image) or the discrete roto-translation group \(p4\) (for which \(|G|\) is height\(\times\)width\(\times4\), since there are \(4\) rotation in \(p4\)).
\section{Experiments}
In this section we describe the dataset used and motivate our baseline architecture. The results are then divided into models that use convolution only and models that use self\-/attention, including ASC. We then present the overall trends. We specify various hyperparameters in the Appendix \ref{sec:hyperparam}.

\textbf{Dataset} In our experiments we test the models' performance on the CIFAR10 and CIFAR100 datasets \citep{CIFAR10}. These consist of \(60k\) images each, \(50k\) for training and \(10k\) for testing. We further split the training set into \(45k\) images for training and we leave \(5k\) for validation. CIFAR10 consists of \(10\) classes and CIFAR100 consists of 100 classes. 

\textbf{Backbone} ResNets \citep{HeZRS15} are a family of CNNs. They are composed of building blocks which are either basic blocks, which have \(2n\) layers per feature map size or bottleneck blocks, which have \(3n\) layers per feature map size. For CIFAR the feature map sizes are \((32, 16, 8)\). Based on the choice of \(n\) \citet{HeZRS15} define the depth of the network. On top of this, they also count the initial and the final layer. As a result we can choose basic block ResNets with \(6n+2\) layers or bottleneck block ResNets with \(9n+2\) layers. In our experiments we use a variant of the ResNet that is appropriate for CIFAR and also uses bottleneck residual blocks. We take this approach because the models using self\-/attention in the literature use it as a replacement for the \(3\times 3\) layer inside the bottleneck residual block and we do the same. As a result, from the standard ResNet20, which is an example of a ResNet with basic residual block (\(6n+2\) layers, with \(n=3\)), we arrive at ResNet29, which is an example of a ResNet with bottleneck residual block (\(9n+2\) layers, with \(n=3\)).

\begin{figure}[h]
\centering
\includegraphics[scale=0.6]{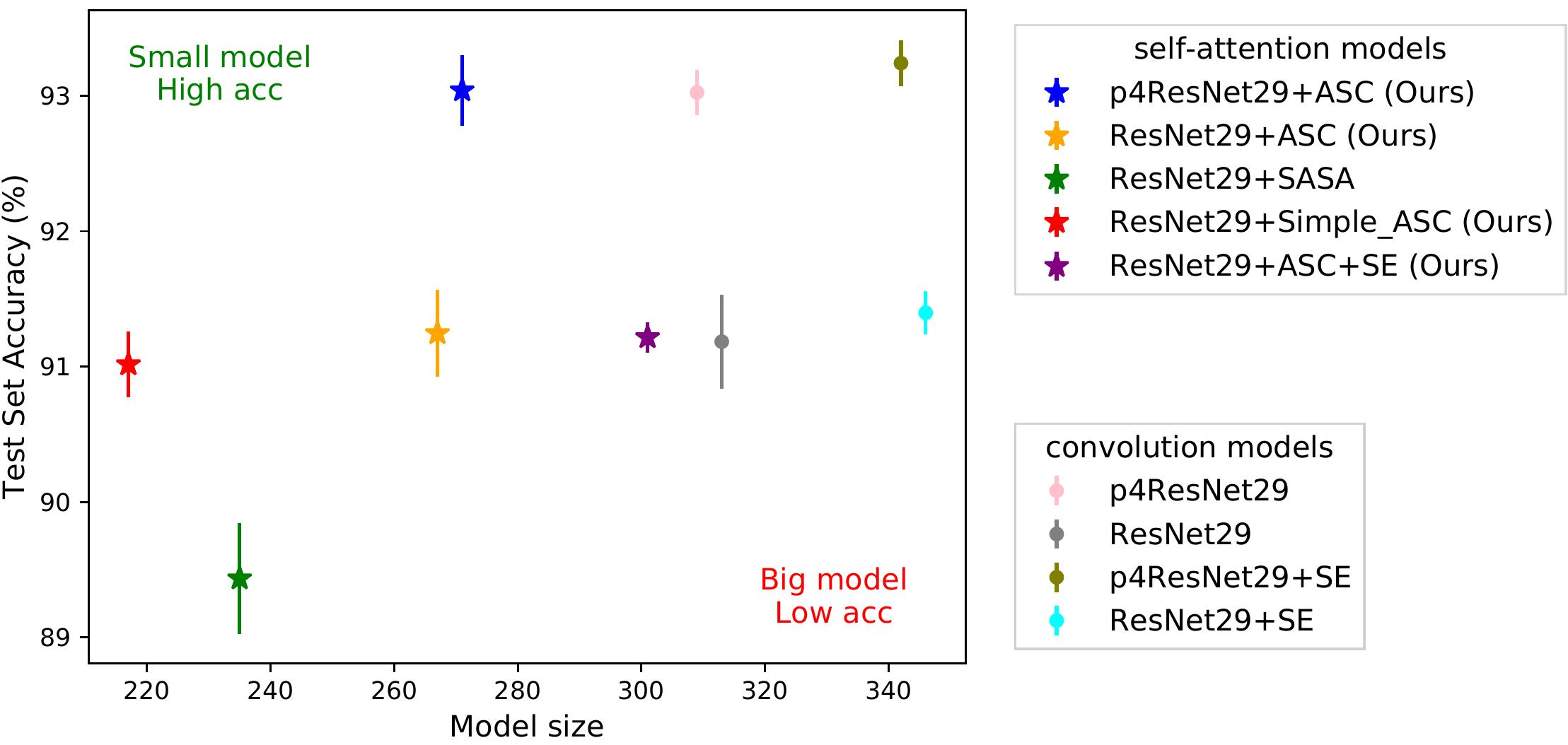}
\caption[Performance of ResNet29 variants on Cifar10]{Comparing the model size (in thousands of parameters) to accuracy of several variants of ResNet29 on CIFAR10. Error bars represent 1 standard deviation from 8 runs. Precise numbers are in Appendix Table \ref{tab:acc_29_cifar10}.} \label{fig:CIFAR10_resnet29}
\end{figure}

\textbf{Models} We present the results on CIFAR10 in Figure \ref{fig:CIFAR10_resnet29}. The models are divided into convolution only models and self\-/attention models, including ASC. Convolution models include squeeze and excite models (SE), since these do not change the convolution operation, but adds the SE module at the end of each bottleneck block. We show the baseline ResNet29, the roto\-/translational baseline \(p4\)ResNet29, together with these models with SE. We now turn to models that use self\-/attention, for which we replace the \(3\times 3\) convolutions in ResNet29s bottleneck with self\-/attention. This means that we still leave a convolution, the convolution in the stem (this is the first layer in all ResNets). We experiment with several models. We replicate the strategy for positional embeddings in \citet{ramachandran2019SASA} and we only learn a factorized variant of \(\beta^K\), instead of learning an affine map for each of the \(Q,K,V\) terms. These parameters are factorized over the spatial dimensions. This means that for each head, half of the positional embeddings \(\beta^K\) is invariant to horizontal translations, while the other half is invariant to vertical translations. Models using these parametrization of self\-/attention are denoted with +SASA. Self\-/attention models includes ASC models. We use Simple\_ASC based on Equation \ref{eqn:Z2_simple_selfattn_expanded}. This does not use three separate pairs of parameters, just one. With +ASC we denote the general form of ASC as described in equation \ref{eqn:Z2_ASC_model}. We also include ResNet29+ASC+SE as one of the models. This is because ASC and SE are not mutually exclusive and we can add the SE module as for a standard ResNet. We also train a roto-translation ASC, which we denote with \(p4\)ResNet29+ASC. This uses the general form of roto\-/translation ASC as described in equation \ref{eqn:p4_ASC_model}.

\textbf{Results} The overall trade-off between accuracy and parameter count in Figure \ref{fig:CIFAR10_resnet29} shows that all the self\-/attention models use fewer parameters than convolution models, while reaching an accuracy in the same range. This confirms the intuition that data dependent filters are more flexible and a powerful modeling choice. We see that +SE adds a small improvement to convolutional models, but no improvement to +ASC models. This indicates that local self\-/attention provides enough spatial context. In terms of the self\-/attention models, +SASA is the only one which does not manage to compete with the baseline. We conclude that a relative affine map is required, additive positional embeddings are not enough. Most insightful, +Simple\_ASC drastically decreases the number of parameters (\(\approx 30\%\)) and reaches accuracy within one standard deviation of the baseline. The three models with the highest accuracy are the roto-translational counterparts to standard models. They show a clear constant increase in performance with little to no extra parameters.
\begin{table}
\centering
\caption[Performance of ResNet29 variants on Cifar10]{We show accuracy mean and 1 standard deviation from 8 runs while varying random seed of various models on CIFAR100. Plot variant in Appendix Figure \ref{fig:CIFAR10_resnet29}} \label{tab:CIFAR100_resnet29}
\begin{tabular}{c|c|c|c}
\hline
& Model & Accuracy & \#Parameters \\
    \hline
    translation & ResNet29 & 68.34 $\pm$ 0.38 & 336$k$ \\
    equivariant & ResNet29+Simple\_ASC (ours) & 68.40 $\pm$ 0.78 & \textbf{240}$\mathbf{k}$ \\
    & ResNet29+ASC (ours) & \textbf{68.68} $\mathbf{\pm}$ \textbf{0.77} & 291$k$ \\
    \hline
    roto-translation & $p4$ResNet29 & 72.03 $\pm$ 0.46 & 321$k$ \\
    equivariant & \(p4\)ResNet29+SE & 72.03 $\pm$ 0.45 & 354$k$ \\
    & \(p4\)ResNet29+ASC (ours) & \textbf{72.71} $\mathbf{\pm}$ \textbf{0.51} & \textbf{283}$\mathbf{k}$ \\
\hline
\end{tabular}
\end{table}This confirms that the theory was applied consistently and that the group theoretic approach benefits attention mechanisms. We ran a subset of the models on Cifar100, results are shown in Table \ref{tab:CIFAR100_resnet29}. The table shows that both ASC and roto-translation equivariance are more beneficial on this dataset. This indicates that these models generalize better when there is less data available.

The experiments show that self\-/attention is competitive when including the affine map as done in ASC and that roto\-/translational equivariance is a robust improvement in all models. 
\section{Related work}
\textbf{Group equivariant CNNs} The theoretically founded approach of group equivariant neural networks has motivated several advances. These works are presented under a unifying framework of group equivariant convolutional networks in \citet{cohen2018general}. Closely related to our work are the developments in planar Euclidean groups in \citet{worrall2017harmonic,weiler2018learning,diaconu2019learning} and the discrete variants of these in \citet{Lecun89,CohenW16,dieleman2016exploiting,hoogeboom2018hexaconv} and 3D in \citet{kondor2018n,cohen2018spherical,esteves2018learning,worrall2018cubenet,weiler20183d,winkels20183d,thomas2018tensor}. We note that this work would benefit from extensions to semigroups \citep{worrall2019deep} or curved manifolds \citep{cohen2019gauge}. Other relevant works include \citet{kondor2018generalization,esteves2019equivariant,esteves2018cross,marcos2017rotation,zhou2017oriented,bekkers2018roto,jacobsen2017dynamic}.

\textbf{Self-attention} Various forms of self\-/attention for CNNs have been introduced based on non-local means \citep{buades2005non} or on the Transformer \citep{Vaswani17}. These models are generally trained for classification/segmentation \citep{wang2018NLNet,hu2019LRNet}, but have also been used for generative tasks \citep{parmar2018imagetransformer,zhang2018SAGAN}. Some works \citep{hu2019LRNet} show that locality of the self\-/attention mechanism together with softmax helps and that relative positional embeddings are essential. In parallel, \citet{ramachandran2019SASA} also show improvements using local self\-/attention with local relative positional embeddings over \citet{bello2019AACNN} which use global self\-/attention with global positional embeddings. We describe in more detail how our model is related to \citet{ramachandran2019SASA,hu2019LRNet} at the end of Section \ref{sec:ASC}. It is also worth mentioning that several of these works compare convolutional models with attention based models and show that convolutional models require more floating point operations per second (FLOPS). Self\-/attention has also been applied to graphs \citep{velivckovic2017graph}. Regardless, they are faster than the attention based models.

\textbf{Data dependent filters} Other works which approach the problem of learning to apply filters differently at each spatial positions are \citet{stanley2019designing,jia2016dynamic,ha2016hypernetworks,jaderberg2015spatial,sabour2017dynamic,kosiorek2019stacked,dai2017deformable,su2019pixel}.
\section{Conclusion}
In this work we show that there is a mixture of convolution and self\-/attention that can be used to replace spatial convolutions in CNNs. This module can be described as a convolution with data dependent filters. By retaining all the benefits of self\-/attention and convolution, what emerges are filters that are translationally equivariant, while being applied differently for each location in the input. The results show that this method is able to achieve comparable if not better performance than the convolutional models, while using fewer parameters and a bigger receptive field. Under simplifying assumptions, we can recover both self\-/attention and convolution, which allows us to incorporate the group theoretic approach of Group equivariant CNNs. Therefore, we prove the translational equivariance of ASC and we also develop the roto-translation equivariant ASC. The latter, is more robust to transformations of the input while surpassing the other models in accuracy. We expect the most fruitful directions for future work to be: an efficient implementation (because self\-/attention is slower), efficient parametrization (order and shape of \(W\)s and \(\mathcal{A}\)s), merge self\-/attention and convolution for NLP/graphs, and equivariant ASC for manifolds (equivariant self\-/attention could score transport methods without assuming a predefined geometry). 
\bibliography{bibliography}
\bibliographystyle{iclr2020_conference}

\newpage
\appendix
\section{QKV ASC Figure}
\begin{figure}[h]
\includegraphics[scale=0.52]{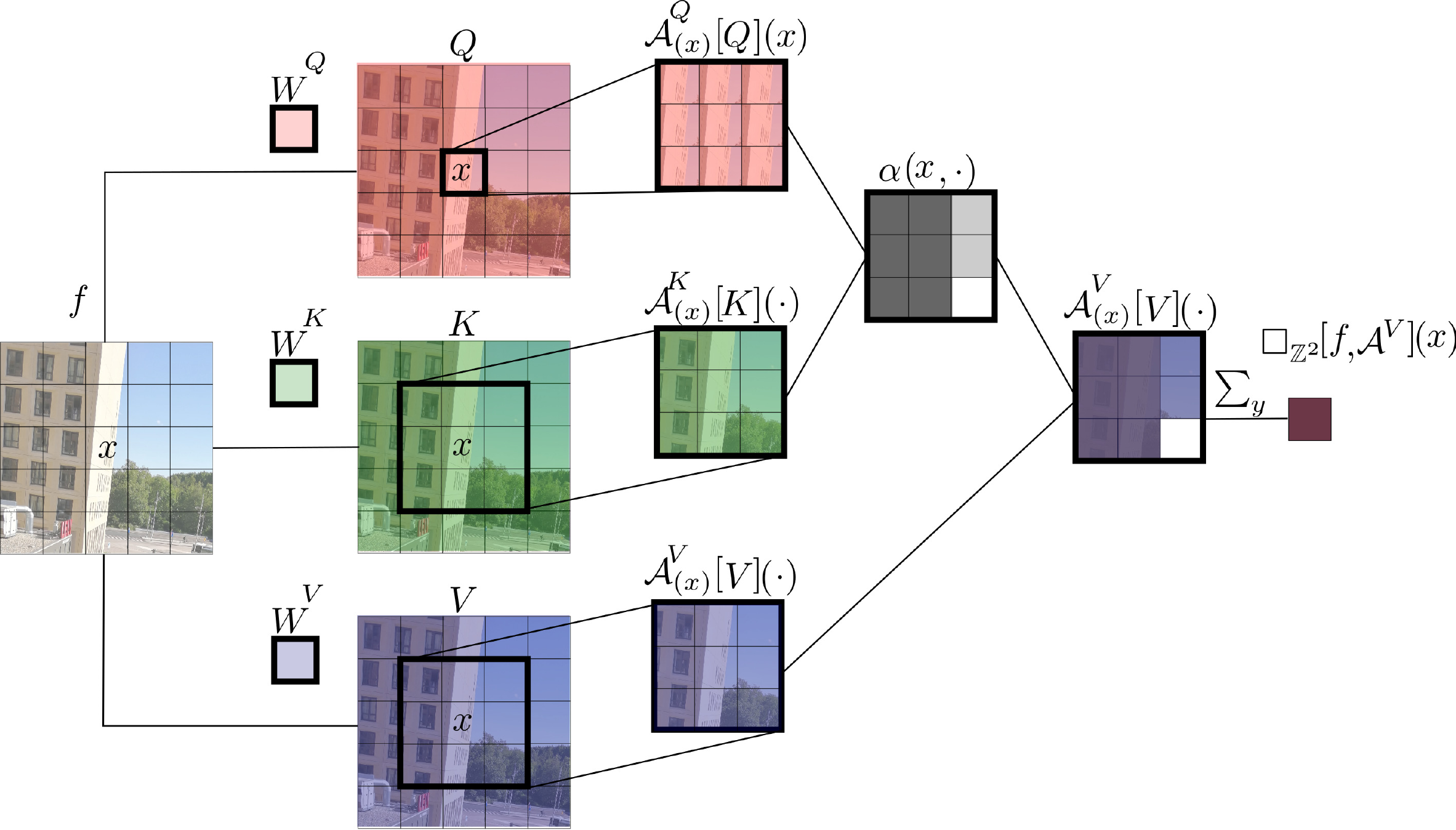}
\centering
\caption[Affine Self Convolution]{Affine Self Convolution with extent \(3\)\(\times\)\(3\) evaluated at \(x\). For clarity, the affine parameters are not depicted, just the output of the affine map. As opposed to Simple ASC in Figure \ref{fig:Z2_Simple_ASC}, here we have 3 linear transformations \(W^Q, W^K, W^V\), which do not process spatial information, just output 3 separate embeddings \(Q,K,V\) for the same input \(f\). Moreover, three local affine maps \(\mathcal{A}^Q, \mathcal{A}^K, \mathcal{A}^V\) are applied to each neighborhood relative to a center \(x\) in order to process spatial information. The score \(\alpha(x,y)\) is evaluated between the center taken from \(Q\) and the neighbors taken from \(K\). The score is then used to aggregate neighbors taken from \(V\). This depicts how self\-/attention is applied in practice.} \label{fig:ASC}
\end{figure}
\section{Results}
\begin{figure}[h]
\centering
\includegraphics[scale=0.6]{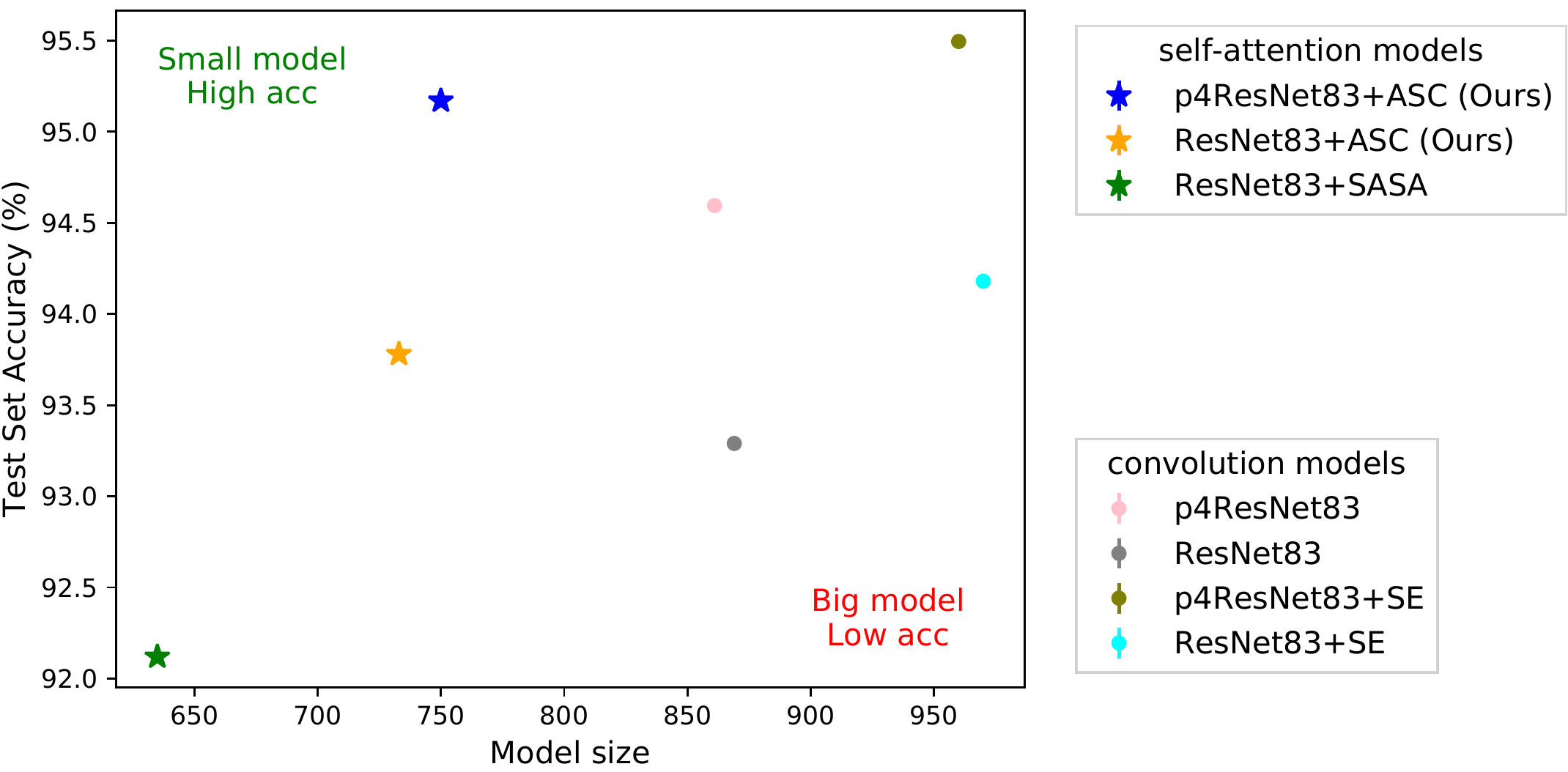}
\caption[Performance of ResNet83 variants on Cifar10]{Comparing the parameter efficiency (model size) to accuracy of several variants of ResNet83 on CIFAR10. Model size is divided by 1000. These results are from only 1 run of each model.}\label{fig:CIFAR10_resnet83}
\end{figure}

\begin{table}[h]
    \centering
    \caption[Performance of ResNet29 variants on Cifar10]{We show accuracy mean and 1 standard deviation from 8 runs while varying random seed of various models on CIFAR10. With \textbf{bold} we outline the highest accuracy second highest with \textcolor{blue}{\textbf{blue}}. We do the same for the most parameter efficient models. Plot variant in Figure \ref{fig:CIFAR10_resnet29}.}
    \begin{tabular}{c|c|c|c}
    & Model & Accuracy & \#Parameters \\
    \hline
    & ResNet29 & 91.18 $\pm$ 0.34 & 313k \\
    convolution& ResNet29 + SE & 91.39 $\pm$	0.15 & 347k \\
    models& \(p4\)ResNet29 & 93.02 $\pm$ 0.16 & 310k \\
    & \(p4\)ResNet29 + SE & \textbf{93.24} $\mathbf{\pm}$ \textbf{0.17} & 342k \\
    \hline
    & ResNet29+SASA	            & 89.43 $\pm$ 0.40 & \textcolor{blue}{\textbf{235}\(\mathbf{k}\)} \\
    & ResNet29+Simple\_ASC (ours)	& 91.01 $\pm$ 0.24 & \textbf{217}\(\mathbf{k}\) \\
    self-attention & ResNet29+ASC (ours)         & 91.24 $\pm$ 0.32 & 268\(k\) \\
    models & ResNet29+ASC+SE (ours)      & 91.21 $\pm$ 0.11 & 301\(k\) \\
    & \(p4\)ResNet29+ASC (ours)   & \textcolor{blue}{\textbf{93.03} $\mathbf{\pm}$ \textbf{0.25}} & 272\(k\) \\
    \end{tabular}
    \label{tab:acc_29_cifar10}
\end{table}
\begin{figure}[h]
\centering
\includegraphics[scale=0.6]{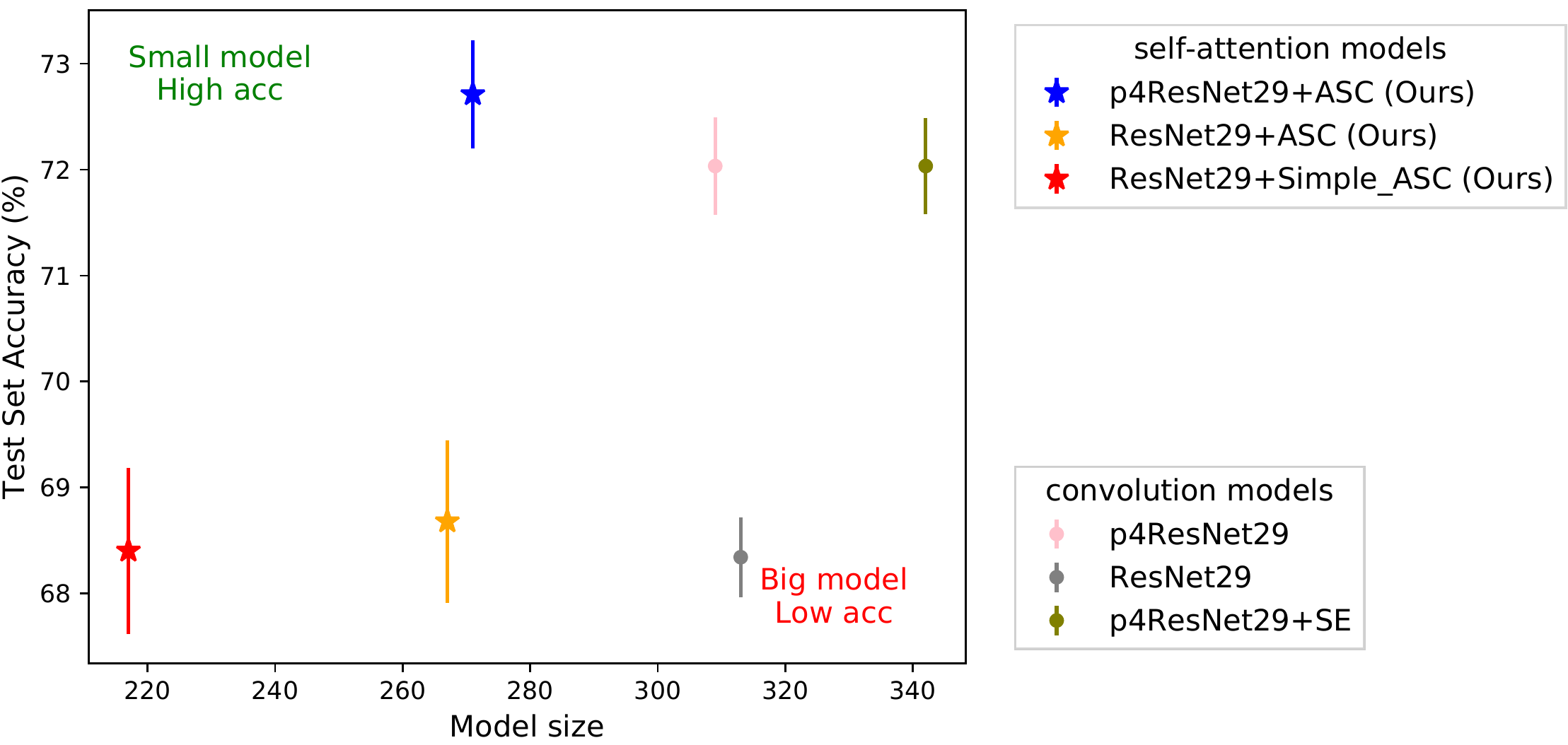}
\caption[Performance of ResNet29 variants on Cifar100]{Comparing the model size (in thousands of parameters) to accuracy of several variants of ResNet29 on CIFAR100. Error bars represent 1 standard deviation from 8 runs. Precise numbers are in Table \ref{tab:CIFAR100_resnet29}.} \label{fig:CIFAR100_resnet29}
\end{figure}
We also run preliminary experiments with ResNet83 (bottleneck block \(9n+2\), with \(n=9\)), which are presented in Figure \ref{fig:CIFAR10_resnet83}. These show a similar trend to the ResNet29 models, but they also indicate that attention, either as SE or ASC might be more rewarding in bigger models. 
\section{Hyperparameters, initialization and training schedule} \label{sec:hyperparam}
We normalize the images and use the standard data augmentation technique of random horizontal flips and random crops of 32\(\times\)32 from the zero padded 36\(\times\)36 images.

We initialize convolutional layers using He initialization \citep{HeInit} (for the \(p4\) variant, the number of channels is multiplied by 4 in the He initialization) and we initialize batchnorms scaling coefficient \(\gamma\) to 1 and shifting coefficient \(\beta\) to 0. The reduction ratio in SE is \(16\), while in \(p4\) models, the reduction ratio in SE is \(4\).

In the self\-/attention models we replace the spatial convolutions in the bottleneck layers with self\-/attention layers. Where the baseline ResNet uses a stride of 2, the self\-/attention models applies self\-/attention then an average pooling layer with kernel size \(2\)\(\times\)\(2\) and stride 2. The self\-/attention layers use the multi head mechanism with 8 heads and a kernel size of \(5\). We set \(d_k=d_{\text{out}}\). In all examples we initialize \(\psi,\beta\sim\mathcal{N}(0,1)\). We use the dot product score normalized by \(\frac{1}{d_k}\). These models are more unstable in the first couple of epochs. Therefore, we initialize the scaling coefficient \(\gamma\) of the last batchnorm in each residual block to 0 as done in \citet{ZeroInit}. Moreover, we warmup (per epoch) the learning rate for the first 10 epochs, up to the learning rate of 0.1. The models are trained using Nesterov accelerated gradient with momentum 0.9 and weight decay of 0.0001. The ResNet29 models and its variants are trained for 100 epochs, where the learning rate was divided by 10 at epochs 50 and 75 and each model was trained for 100 epochs. We also include examples of ResNet83. These bigger models we trained for 200 epochs and divided the learning rate by 10 at epochs 100 and 150. We do this for all the models, for a fair comparison. Throughout our experiments we used Pytorch \citep{paszke2017pytorch}. We have also taken inspiration from the GitHub repositories: \citet{imgclsmob,propercifar10resnet}.

When we use roto-translation layers instead of standard layers, we divide the number of channels in each stage of the ResNet by \(\sqrt{|p4|}=2\). This preserves a similar number of parameters between the roto-translation models and the standard models.
\section{Roto-translation Simple ASC Figure}
\begin{figure}[h]
\includegraphics[scale=0.52]{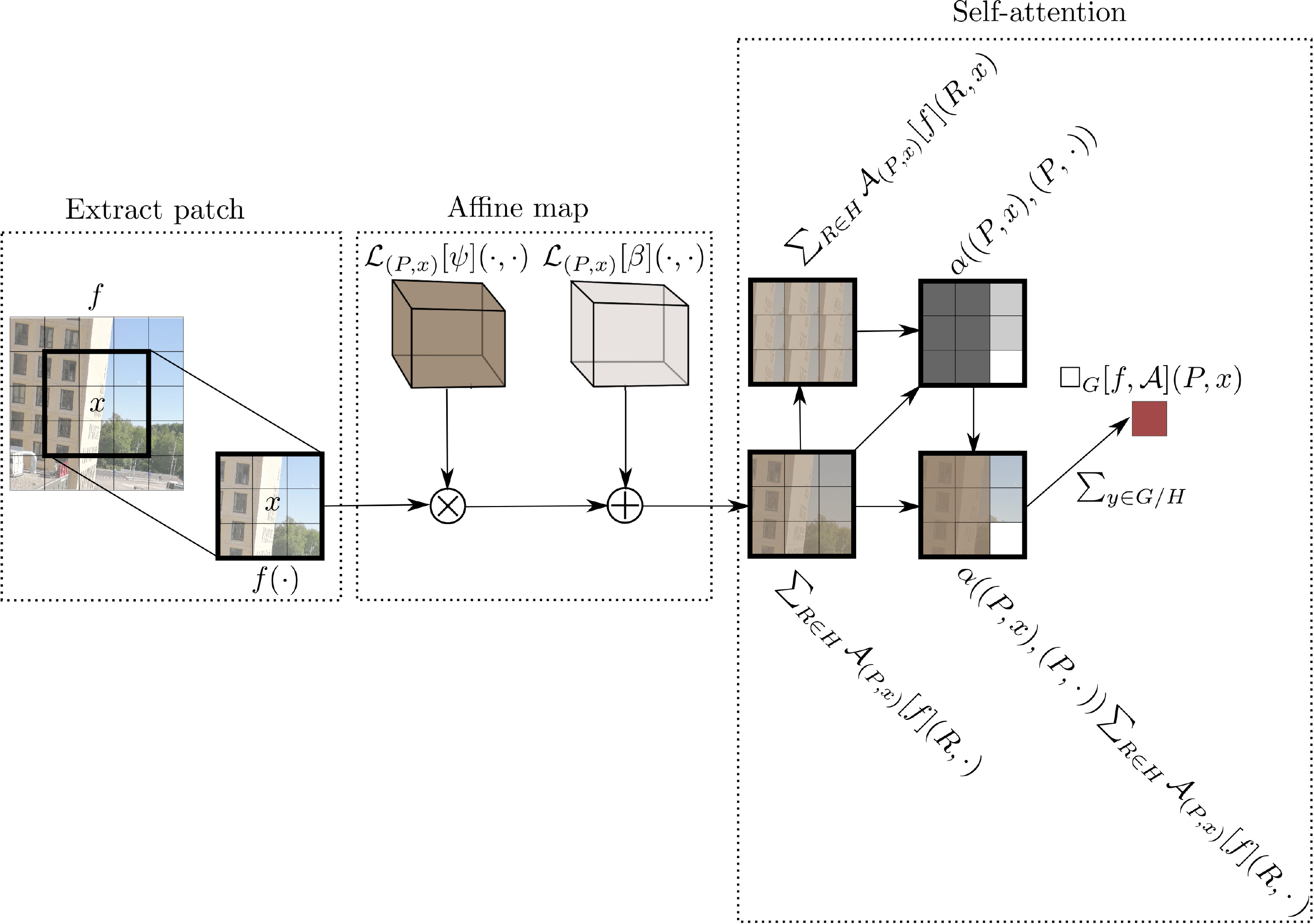}
\centering
\caption[Roto-translation Simple ASC]{Roto-translation Simple ASC for an input \(f\) and an affine map \(\mathcal{A}\) with extent \(3\)\(\times\)\(3\) evaluated at \((P,x)\in G\). Similarly to the planar case, \((\cdot)\) denotes the neighborhood of \(x\) for \(f(\cdot)\). All the other functions depicted have roto\-/translational domain. For these, \(\cdot\) as the second argument, \((,\cdot)\) also represents the neighborhood of \(x\), but, as the first argument, \((\cdot,)\) represents all the rotation values in the group \(G\). As a result, the filter \(\psi\) is a three dimensional filter, with rotations and translations. Because of the score function we chose, the self\-/attention mechanism is evaluated for each \(P\) independently. Therefore, the self\-/attention looks very similar to the depiction in Figure \ref{fig:Z2_Simple_ASC}.} \label{fig:Group_Simple_ASC}
\end{figure}
\section{Roto-translation ASC} \label{sec:p4_ASC}
In practice, the input is discrete and we turn to functions on \(p4\), \(G\coloneqq p4, H\coloneqq C_4, \mathbb{Z}^2=G/H\). For the general roto-translation ASC all quantities are defined analogously to the ASC on \(\mathbb{Z}^2\) in Equation \ref{eqn:Z2_ASC_model}, using the score function in Equation \ref{eqn:p4_score_fn} and replacing the domain \(\mathbb{Z}^2\) with \(G\). Therefore, we define roto-translation ASC with:
\begin{align}
\shortintertext{1) three linear mappings of the input \(f\) (defined analogously for \(K\) and \(V\)):}
    Q_c(P,x) =& [f\star_{G}W^Q_c](P,x)
\shortintertext{2) three affine maps \(\mathcal{A}^Q, \mathcal{A}^K, \mathcal{A}^V\) for the \(Q,K,V\) terms (defined analogously for \(K\) and \(V\)):}
    \mathcal{A}^Q_{(P,x)}[Q]_c(R,y) =& Q_c(R,y)\mathcal{L}_{(P,x)}\psi^Q_c(R,y)+\mathcal{L}_{(P,x)}\beta^Q_c(y) \\
\shortintertext{3) a score function between center \((P,x)\) and neighbor \(y\), which is then normalized with softmax:}
    \begin{split}
    \alpha((P,x),(P,y)) =& \text{softmax}_y(\text{score}\left(\left ( \sum_{R\in H} \mathcal{A}^Q_{(P,x)}[Q](R,x)\right )\right.,\left.\left ( \sum_{R\in H} \mathcal{A}^K_{(P,x)}[K](R,y)\right )\right))
    \end{split}
\shortintertext{4) an aggregation of the \(V\) embeddings based on the score function:
The final aggregation step completely describes the roto-translation ASC operation:}
    \square_G[V,\mathcal{A}^V]_c(P,x) &= \sum_{y\in\mathbb{Z}^2} \alpha((P,x),(P,y))\sum_{R\in H}\mathcal{A}_{(P,x)}^V[V]_c(R,y) \label{eqn:p4_ASC_model}
\end{align}Similarly to ASC, for roto-translation ASC we learn \(\psi^Q, \beta^Q\) for only one spatial index \(0\).

\section{Proofs}
\begin{tcolorbox}[breakable, 
boxrule=0.5mm,
coltitle=black,
colframe=lightgray,
colback=white,
width=(1\linewidth),
adjusted title={ASC translation equivariance proof}]

\allowdisplaybreaks
Claim:
\begin{align}
    \square_{\mathbb{Z}^2}[\mathcal{L}_{(z)}[f],\mathcal{A}](x) =& \mathcal{L}_{(z)}[\square_{\mathbb{Z}^2}[f,\mathcal{A}]](x) \label{eqn:Z2_ASC_proof}
\end{align}
Proof:
\begin{align}
    \shortintertext{Expanding from the left hand side:}
    \begin{split}
    \square_{\mathbb{Z}^2}[\mathcal{L}_{(z)}[f], \mathcal{A}](x) =& \sum_{y\in\mathbb{Z}^2}\text{softmax}_y(\text{score}(\mathcal{A}_{(x)}[\mathcal{L}_{(z)}[f]](x), \mathcal{A}_{(x)}[\mathcal{L}_{(z)}[f]](y))) \nonumber\\ &\mathcal{A}_{(x)}[\mathcal{L}_{(z)}[f]](y)
    \end{split}
    \shortintertext{Using the substitution: \(y \mapsto y+z\)}
    \shortintertext{Using: \(y \mapsto y+z \Rightarrow \mathcal{A}_{(x)}[\mathcal{L}_{(z)}[f]](y) = \mathcal{A}_{(-z+x)}[f](y)\), which we prove in the Appendix \ref{eqn:Z2_ASC_proof1}.}
    \shortintertext{Using: \( \mathcal{A}_{(x)}[\mathcal{L}_{(z)}[f]](x) = \mathcal{A}_{(-z+x)}[f](-z+x)\), which we prove in the Appendix \ref{eqn:Z2_ASC_proof2}.}
    \begin{split}
        =& \sum_{y\in\mathbb{Z}^2}\text{softmax}_y(\text{score}(\mathcal{A}_{(-z+x)}[f](-z+x), \mathcal{A}_{(-z+x)}[f](y))) \\ &\mathcal{A}_{(-z+x)}[f](y)
    \end{split} \nonumber
    \\
    =& \square_{\mathbb{Z}^2}[f,\mathcal{A}](-z+x) \nonumber \\
    =& \mathcal{L}_{(z)}[\square_{\mathbb{Z}^2}[f,\mathcal{A}]](x) \nonumber
\end{align} 
\end{tcolorbox}

\begin{tcolorbox}[
breakable,
boxrule=0.5mm,
coltitle=black,
colframe=lightgray,
colback=white,
width=(1\linewidth),
adjusted title={Proof }]
\allowdisplaybreaks
\begin{align}
    \shortintertext{Claim:}
    y \mapsto y+z &\Rightarrow \mathcal{A}_{(x)}[\mathcal{L}_{(z)}[f]](y) = \mathcal{A}_{(-z+x)}[f](y) \label{eqn:Z2_ASC_proof1}
    \shortintertext{Proof:}
    \begin{split}
    \mathcal{A}_{(x)}[\mathcal{L}_{(z)}[f]](y) =& \mathcal{L}_{(z)}[f](y)\psi(y-x)+\beta(y-x) \\
    =& f(y-z)\psi(y-x)+\beta(y-x) 
    \shortintertext{ Using the substitution: \(y \mapsto y+z\)}
    =& f(y)\psi(y+z-x)+\beta(y+z-x) \\
    =&\mathcal{A}_{(-z+x)}[f](y) 
    \end{split} \nonumber
\end{align}
\end{tcolorbox}

\begin{tcolorbox}[
breakable,
boxrule=0.5mm,
coltitle=black,
colframe=lightgray,
colback=white,
width=(1\linewidth),
adjusted title={Proof }]
\allowdisplaybreaks
\begin{align}
    \shortintertext{Claim:}
    \mathcal{A}_{(x)}[\mathcal{L}_{(z)}[f]](x) &= \mathcal{A}_{(-z+x)}[f](x-z)  \label{eqn:Z2_ASC_proof2}
    \shortintertext{Proof:}
    \begin{split}
    \mathcal{A}_{(x)}[\mathcal{L}_{(z)}[f]](x) =&  \mathcal{L}_{(z)}[f](x)\psi(x-x)+\beta(x-x) \\
    =& f(x-z)\psi((x-z)-(x-z))+\beta((x-z)-(x-z)) \\
    =&\mathcal{A}_{(x-z)}[f](x-z)
    \end{split} \nonumber
\end{align}
\end{tcolorbox}

\begin{tcolorbox}[
breakable,
boxrule=0.5mm,
coltitle=black,
colframe=lightgray,
colback=white,
width=(1\linewidth),
adjusted title={Proof \(\beta\) is a function on \(G/H\)}]
Claim:
\begin{align}
    \sum_{R\in H}\mathcal{A}_{(P,x)}[f](R,y) &= \sum_{R\in H} f(R,y)\mathcal{L}_{(P,x)}[\psi](R,y)+\mathcal{L}_{(P,x)}[\beta](y) \label{eqn:p4_affine_map_Z2_beta}
\end{align}
Proof: \\
For the ASC on groups, the map \(\mathcal{A}\) and therefore, \(\alpha\) and \(\beta\), are always used inside \(\sum_{R\in H}\). This leads to a more parameter efficient parametrization for \(\beta\):
\begin{align}
    \begin{split}
    \sum_{R\in H}\mathcal{A}_{(P,x)}[f](R,y) &= \sum_{R\in H} f(R,y)\mathcal{L}_{(P,x)}[\psi](R,y)+\mathcal{L}_{(P,x)}[\beta](R,y) \\
    &= \sum_{R\in H} f(R,y)\mathcal{L}_{(P,x)}[\psi](R,y)+\sum_{R\in H}\mathcal{L}_{(P,x)}[\beta](R,y) \\
    \sum_{R\in H}\mathcal{L}_{(P,x)}[\beta](R,y) &= \sum_{R\in H}\beta(P^{-1}R,P^{-1}(y-x))
    \shortintertext{Using the substitution: \(R\mapsto PR\)}
    &=\sum_{R\in H}\beta(1,P^{-1}(y-x))
    \end{split} \nonumber
\end{align}

This is actually a function on \(G/H\), not the whole group \(G\). Therefore, we replace \(\sum_{R\in H}\beta(R,x)\) with \(\beta(x)\).
\end{tcolorbox}

\begin{tcolorbox}[
breakable,
boxrule=0.5mm,
coltitle=black,
colframe=lightgray,
colback=white,
width=(1\linewidth),
adjusted title={Roto-translation ASC equivariance proof}]
Claim:
\begin{align}
\square_G[\mathcal{L}_{(S,z)}[f],\mathcal{A}](P,x)=\mathcal{L}_{(S,z)}[\square_G[f,\mathcal{A}]](P,x) \label{eqn:p4_ASC_equivariance} 
\end{align}
Proof:
\allowdisplaybreaks
\begin{align}
    \shortintertext{Expanding on the left hand side:}
    \allowdisplaybreaks
    \begin{split}
    \square_G[\mathcal{L}_{(S,z)}[f],\mathcal{A}](P,x) =&\sum_{y\in G/H} \text{softmax}_y \left ( \left ( \sum_{R\in H} \mathcal{A}_{(P,x)}[\mathcal{L}_{(S,z)}[f]](R,x)\right )  \right.\\ 
    &\left. \left ( \sum_{R\in H} \mathcal{A}_{(P,x)}[\mathcal{L}_{(S,z)}[f]](R,y)\right ) \right) \\ 
    &\left ( \sum_{R\in H} \mathcal{A}_{(P,x)}[\mathcal{L}_{(S,z)}[f]](R,y)\right )
    \end{split} \nonumber
    \shortintertext{Using the substitution: \(R\mapsto SR\) and \(y\mapsto Sy+z\)}
    \shortintertext{Using: \(R\mapsto SR \Rightarrow \mathcal{A}_{(P,x)}[\mathcal{L}_{(S,z)}[f]](R,x)=\mathcal{A}_{(S,z)^{-1}(P,x)}[f](R,S^{-1}(x-z))\), which we prove in the Appendix \ref{eqn:p4_ASC_proof1}.}
    \shortintertext{Using: \(R\mapsto SR, y\mapsto Sy+z \Rightarrow \mathcal{A}_{(P,x)}[\mathcal{L}_{(S,z)}[f]](R,y)=\mathcal{A}_{(S,z)^{-1}(P,x)}[f](R,y)\), which we prove in the Appendix \ref{eqn:p4_ASC_proof2}.}
    \allowdisplaybreaks
    \begin{split}
    =&\sum_{y\in G/H} \text{softmax}_y \left ( \left ( \sum_{R\in H} \mathcal{A}_{(S,z)^{-1}(P,x)}[f](R,S^{-1}(x-z))\right )  \right.\\ 
    &\left. \left ( \sum_{R\in H} \mathcal{A}_{(S,z)^{-1}(P,x)}[f](R,y)\right ) \right)  \\ 
    &\left ( \sum_{R\in H} \mathcal{A}_{(S,z)^{-1}(P,x)}[f](R,y)\right )
    \end{split} \nonumber
    \\[2ex]
    =&\square_G[f,\mathcal{A}](S^{-1}P,S^{-1}(x-z))  \nonumber\\
    =&\mathcal{L}_{(S,z)}[\square_G[f,\mathcal{A}]](P,x)  \nonumber
\end{align}
By arriving at the right hand side of equation \ref{eqn:p4_ASC_equivariance}, we concolude the proof that roto-translation ASC is equivariant to actions of the group \(SE(2)\).
\end{tcolorbox}

\begin{tcolorbox}[
breakable,
boxrule=0.5mm,
coltitle=black,
colframe=lightgray,
colback=white,
width=(1\linewidth),
adjusted title={Proof }]
\allowdisplaybreaks
\begin{align}
    \shortintertext{Claim:}
    R\mapsto SR &\Rightarrow \mathcal{A}_{(P,x)}[\mathcal{L}_{(S,z)}[f]](R,x)=\mathcal{A}_{(S,z)^{-1}(P,x)}[f](R,S^{-1}(x-z)) \nonumber \\
    \label{eqn:p4_ASC_proof1}
    \shortintertext{Proof:}
    \begin{split}
    \mathcal{A}_{(P,x)}[\mathcal{L}_{(S,z)}[f]](R,x) =& \mathcal{L}_{(S,z)}[f](R,x)\mathcal{L}_{(P,x)}[\psi](R,x)+\mathcal{L}_{(P,x)}[\beta](x) \\
    =& f(S^{-1}R,S^{-1}(x-z))\psi(P^{-1}R,P^{-1}(x-x))+\beta(P^{-1}(x-x))
    \shortintertext{Using the substitution: \(R\mapsto SR\)}
    =& f(R,S^{-1}(x-z))\psi(P^{-1}SR,P^{-1}(x-x))+ \\ &+\beta(P^{-1}(x-x)) \\
    =& f(R,S^{-1}(x-z))\psi(P^{-1}SR,P^{-1}S(S^{-1}(x-z)-S^{-1}(x-z)))+\\&+\beta(P^{-1}S(S^{-1}(x-z)-S^{-1}(x-z))) \\
    =& f(R,S^{-1}(x-z))\mathcal{L}_{(S^{-1}P,S^{-1}(x-z))}[\psi](R,S^{-1}(x-z))+\\&+\mathcal{L}_{(S^{-1}P,S^{-1}(x-z))}[\beta](S^{-1}(x-z))) \\
    =&\mathcal{A}_{(S^{-1}P,S^{-1}(x-z))}[f](R,S^{-1}(x-z)) \\
    =&\mathcal{A}_{(S,z)^{-1}(P,x)}[f](R,S^{-1}(x-z))
    \end{split} \nonumber
\end{align}
\end{tcolorbox}

\begin{tcolorbox}[
boxrule=0.5mm,
coltitle=black,
colframe=lightgray,
colback=white,
width=(1\linewidth),
adjusted title={Proof }]
\allowdisplaybreaks
\begin{align}
    \shortintertext{Claim:} 
    R\mapsto SR, y\mapsto Sy+z &\Rightarrow \mathcal{A}_{(P,x)}[\mathcal{L}_{(S,z)}[f]](R,y)=\mathcal{A}_{(S,z)^{-1}(P,x)}[f](R,y) \nonumber \\
    \label{eqn:p4_ASC_proof2}
    \shortintertext{Proof:}
    \begin{split}
    \mathcal{A}_{(P,x)}[\mathcal{L}_{(S,z)}[f]](R,y) =& \mathcal{L}_{(S,z)}[f](R,y)\mathcal{L}_{(P,x)}[\psi](R,y)+\mathcal{L}_{(P,x)}[\beta](y) \\
    =& f(S^{-1}R,S^{-1}(y-z))\psi(P^{-1}R,P^{-1}(y-x))+\beta(P^{-1}(y-x))
    \shortintertext{Using the substitutions: \(R\mapsto SR\) and \(y\mapsto Sy+z\)}
    =& f(R,y)\psi(P^{-1}SR,P^{-1}(Sy+z-x))+ \\ &+\beta(P^{-1}(Sy+z-x)) \\
    =& f(R,y)\psi(P^{-1}SR,P^{-1}S(y-S^{-1}(x-z)))+ \\ &+\beta(P^{-1}S(y-S^{-1}(x-z))) \\
    =& f(R,y)\mathcal{L}_{(S^{-1}P,S^{-1}(x-z))}[\psi](R,y)+\\&+\mathcal{L}_{(S^{-1}P,S^{-1}(x-z))}[\beta](y)) \\
    =&\mathcal{A}_{(S^{-1}P,S^{-1}(x-z))}[f](R,y) \\
    =&\mathcal{A}_{(S,z)^{-1}(P,x)}[f](R,y)
    \end{split} \nonumber
\end{align}
\end{tcolorbox}

\begin{tcolorbox}[
boxrule=0.5mm,
coltitle=black,
colframe=lightgray,
colback=white,
width=(1\linewidth),
adjusted title={Proof Group Squeeze is invariant to transformations of the group}]
Claim:
\begin{align}
    \text{squeeze}(\mathcal{L}_{(s)}[f])=\text{squeeze}(f) \label{eqn:squeeze_g_invariant}
\end{align}
Proof:
\begin{align}
    \text{squeeze}(\mathcal{L}_{(s)}[f]) &= \sigma \left (W_1 \left( \text{ReLU} \left ( W_2 \left (\frac{1}{|G|}\sum_{h \in G}\mathcal{L}_{(s)}[f](h) \right ) \right )\right)\right)  \nonumber \\
    &= \sigma \left (W_1 \left( \text{ReLU} \left ( W_2 \left (\frac{1}{|G|}\sum_{h \in G}f(s^{-1}h) \right ) \right )\right)\right)  \nonumber \\
    \shortintertext{Using the substitution: \(h\mapsto sh\)}
    &= \sigma \left (W_1 \left( \text{ReLU} \left ( W_2 \left (\frac{1}{|G|}\sum_{h \in G}f(h) \right ) \right )\right)\right)  \nonumber \\
    &= \text{squeeze}(f) \nonumber
\end{align}
\end{tcolorbox}
\end{document}

%% file: math_commands.tex
\usepackage{amsmath,amsfonts,bm}

\def\eqref#1{equation~\ref{#1}}

\def\1{\bm{1}}

\DeclareMathAlphabet{\mathsfit}{\encodingdefault}{\sfdefault}{m}{sl}
\SetMathAlphabet{\mathsfit}{bold}{\encodingdefault}{\sfdefault}{bx}{n}

